\pdfoutput=1 
\documentclass[sigconf]{acmart}

\AtBeginDocument{%
  \providecommand\BibTeX{{%
    \normalfont B\kern-0.5em{\scshape i\kern-0.25em b}\kern-0.8em\TeX}}}

\setcopyright{acmcopyright}
\copyrightyear{2020}
\acmYear{2020}

\acmConference[KDD '2020]{KDD 2020: ACM Knolwedge Discovery in Databases}{August 23--27, 2020}{San Diego, California}
\acmBooktitle{KDD2020: ACM Knowledge Discovery in Databases,
  August 23--27, 2020, San Diego, California}

\begin{document}

\title{Coronavirus Knowledge Graph: A Case Study}



\author{Chongyan Chen}
\orcid{0000-0003-3188-6652}
\affiliation{%
  \institution{University of Texas at Austin}
  \streetaddress{Guadalupe Street}
  \city{Austin}
  \state{Texas}
  \postcode{78705}
}
\email{chongyanchen_hci@utexas.edu}

\author{Islam Akef Ebeid}
\affiliation{\institution{University of Texas at Austin}}
\email{iaebeid@utexas.edu}

\author{Yi Bu}
\affiliation{%
  \institution{Peking University}
  \city{Beijing}
  \country{China}
}
\email{buyi@pku.edu.cn}



\author{Ying Ding}
\affiliation{\institution{University of Texas at Austin}}
\email{ying.ding@ischool.utexas.edu}


\begin{abstract}


The emergence of the novel COVID-19 pandemic has had a significant impact on global healthcare and the economy over the past few months. The virus's rapid widespread has led to a proliferation in biomedical research addressing the pandemic and its related topics. One of the essential Knowledge Discovery tools that could help the biomedical research community understand and eventually find a cure for COVID-19 are Knowledge Graphs. The CORD-19 dataset is a collection of publicly available full-text research articles that have been recently published on COVID-19 and coronavirus topics. Here, we use several Machine Learning, Deep Learning, and Knowledge Graph construction and mining techniques to formalize and extract insights from the PubMed dataset presented in \cite{dernoncourt2017pubmed} and the CORD-19 dataset \cite{cord2020} to identify COVID-19 related experts and bio-entities. Besides, we suggest possible techniques to predict related diseases, drug candidates, gene, gene mutations, and related compounds as part of a systematic effort to apply Knowledge Discovery methods to help biomedical researchers tackle the pandemic. 

\end{abstract}

\ccsdesc[500]{Applied computing}
\ccsdesc[500]{Applied computing~Health informatics}
\ccsdesc[500]{Applied computing~Biological networks}
\ccsdesc[500]{Computing methodologies~Information extraction}
\ccsdesc[500]{Computing methodologies~Knowledge representation and reasoning}
\ccsdesc[100]{Information systems~Information systems applications}
\keywords{corona virus, named entity Recognition, BioBERT, knowledge Graph, drug discovery}


\maketitle

\section{Introduction}



\subsection{Knowledge Graphs}

A Knowledge Graph (KG) is a graph-based data structure used to represent unstructured information so that a machine can read it. Emerging from Knowledge Bases, KGs now represent a ubiquitous set of methods for representing and integrating knowledge in various domains. A KG contains descriptions of entities and their relationships as information in the form of first-order logical facts such as <Mount Fuji is located in Japan> that can be retrieved and queried heuristically. KGs emerged to power what was known in the eighties and the nineties as Expert Systems - an early form of Artificial Intelligence and Decision Support Systems \cite{russell2016artificial}. The number of entity types and relationships in a KG is finite and is usually but not necessarily organized in a schema or an ontology. In 2012 Google introduced the Google Knowledge Graph \cite{paulheim2017knowledge}, a technology that converts multiple information sources to a graph structure where the nodes represent real-life entities and types. The edges represent the relationships between those entities and types, and the technology was aimed at enhancing the users' search experience through predicting the users' search intent and introducing a Knowledge Panel on the right of page \cite{paulheim2017knowledge}. Knowledge Graph technology today has been adopted in many domains and fields to store, integrate, and represent unstructured information in a structured format that is more flexible and machine-readable than the traditional entity-relationship data model \cite{chen1976entity}. 

Other examples of widely used and adopted KGs in different domains such as Social Networks and Life Sciences include the Facebook Social Graph \cite{ugander2011anatomy} and Chem2Bio2RDF \cite{chen2010chem2bio2rdf}. The advancement of KG construction and mining was powered by the already established research fields of Machine Learning, Deep Learning, Graph Mining, and Complex Networks. The techniques and methods developed by researchers in such fields are used to mine data and information in KGs to extract insights crucial to advancing knowledge in various domains. Research in Information Networks also played a role in the construction and mining of KGs mainly through relying on statistical methods and machine learning techniques \cite{sun2012mining}.  Information networks are widely heterogeneous graphs of nodes and edges representing meta-information about a published corpus of literature such as authors, papers, publications, and venues. Hence, information networks are KGs in which graph mining techniques can be applied to extract insights about author collaboration patterns and their topics of interest. Mining information networks as KGs has to lead to understanding trends such as collaboration patterns and potential drug re-purposing opportunities in a specific domain without reading the entire literature in a field.

KGs have been curated manually, yet, over the years, KG construction techniques have changed. For example, Cyc \cite{lenat1995cyc} is a KG that was manually curated, while Freebase \cite{bollacker2008freebase} and Wikidata \cite{vrandevcic2014wikidata} were crowd-sourced. KGs can also be extracted using Natural Language Processing (NLP) techniques such as in DBpedia \cite{lehmann2015dbpedia} and YAGO \cite{suchanek2007yago}. Alternatively, KGs can be constructed using a combination of manual curation and automatic extraction like in NELL \cite{carlson2010toward} and Knowledge Vault \cite{dong2014knowledge}. Regardless of the approach of how a KG has been constructed, KGs need to be queried and mined to map complex real-world phenomena and eventually be exploited to solve important research questions. For example, Facebook's Social Graph needs to be mined to suggest new friends for users. Life Sciences KGs like Chem2Bio2RDF need to be mined to answer research questions related to biomedical science.

\subsection{Named Entity Recognition in Knowledge Graph Construction}

The move towards natural language understanding through semantic technologies has gained much ground in the past decade, promoting Named Entity Recognition (NER) to a central NLP task. NER has been crucial for building and constructing KGs as the primary method of extracting entities and possibly relations from free text. Also, tasks such as link prediction, relation extraction, and graph completion on KGs are aided by NER. NER can be impactful when applied to mine domain-specific scientific literature such as the biomedical literature to extract bio entities aiding in constructing KGs and advancing downstream knowledge discovery tasks in biomedicine.

Although research in NER has been advancing since the nineties \cite{nadeau2007survey}, early efforts in domain-specific biomedical NER came later in the early 2000s \cite{settles2004biomedical}. Those methods in biomedical NER relied on feature engineering and graphical models such as Hidden Markov Models (HMM) and Conditional Random Fields (CRF) \cite{settles2004biomedical}. When applying CRF models to the biomedical text, the objective is to construct a chain out of the words then predict the assigned labels based on a conditionally trained finite state machine where the probability of each label assigned to a word is correlated with a feature set. The objective was then to maximize the log-likelihood of the label given the word directly. The accuracy of the recognition of bioentities in CRF and HMM models were quite low when compared to state of the art today. The current state of the art relies on the latest in Deep Learning in contextual embedding such as BERT. BERT is a deep learning model developed in \cite{devlin2018bert} by a team at Google to be fine-tuned for machine translation tasks. The model was based on the transformer architecture described in \cite{vaswani2017attention}. Multiple attention heads are used to train a contextual embedding where the task is to predict masked words of the input sentences. The sophisticated inner architecture of BERT based on multiple encoder-decoder layers allows for learning high-quality embedding from a large corpus of data where the learned weights can be later transferred and fine-tuned to downstream tasks. In \cite{lee2020biobert}, the authors trained a BERT model on the corpus of PubMed and PMC named BioBERT. The result was a biomedical contextual embedding model that was later fine-tuned and used in a biomedical NER task producing high accuracy tagging and extraction of bio entities such as drugs, diseases, and genes. The high accuracy of the BioBERT model allowed and aided in the construction of the PubMed KG presented in \cite{xu2020building}.

\subsection{The COVID-19 Knowledge Graph}

Several months amid the emergence of the acute respiratory syndrome COVID-19 caused by the novel coronavirus Sars-CoV-2 in China, the disease has risen to a global pandemic level affecting almost every country on earth and infecting more than 6 million people across the globe and killing more than 350000 \cite{jhu2020}.  As a result, researchers from every domain have reoriented their efforts towards finding ways and solutions to tackle the pandemic. Specifically, the biomedical literature on COVID-19 and SARS-CoV-2 and other related acute respiratory syndromes that have reached an epidemic level such as SARS and MERS have increased exponentially since the virus's appearance back in December 2019. As a result, government-backed calls and research institutions like the Allen Institute for AI have released a COVID-19 Open Research Dataset CORD-19 \cite{cord2020}. The dataset contains over 65000 full-text scholarly articles related to COVID-19, SARS-CoV-2, and other related topics. This effort aims to encourage the NLP and KG researcher community to mine the dataset to generate insights through text mining techniques and methods to help point biomedical researchers in the right direction to fight against the virus. 

We see the release of the CORD-19 dataset of machine-readable scientific literature as an opportunity to extract a comprehensive and cohesive COVID-19 KG of the entities and relationships though cooccurrence within the corpus of articles. The extracted KG will help understand the relationships between the diseases, the genes, the viruses, and the cures involved in and related to COVID-19 so that future graph and network mining efforts can be applied to extract insights from the dataset. Here we present our vision in contributing to that effort. 

We demonstrate methods of entity extraction and KG building to harvest a COVID KG capable of being a useful dataset for future mining in the hope that it will help biomedical researchers find a cure and tackle the pandemic through generating deep insights. We first introduce how to use BioBERT for named entity recognition in the PubMed and CORD-19 datasets. Then we built several Coronavirus Knowledge Graphs based on two different kinds of measurements. One measure the relationship between source node and each target node based on co-occurrence frequency. The other is to use Cosine Similarity to measure the similarity between the source node and each target node.

\section{Related Work}



Previous efforts and trials to build a comprehensive COVID-19 KG have lacked in several areas. For example \cite{domingo2020covid} built a COVID-19 related KG based on 145 articles and provided a web application for ease of use and access. This COVID-19 KG contains 3954 nodes and 9484 relations, covering ten entity types. It reveals host-pathogen interactions, comorbidities, symptoms, and discovered over 300 candidate drugs for COVID-19. Nevertheless, the effort was limited in terms of the number of publications included in constructing the KG. \cite{ge2020data} applied a machine learning model (BERE \cite{hong2019bere}) to integrate and mine KG to also aide in the effort of identifying candidate drugs for COVID-19. Besides, \cite{richardson2020baricitinib} used a pre-built KG for COVID-19 drug discovery and identified the drug "baricitinib" to protect lung cells from being infected by the virus. Previously mentioned efforts though promising, yet they lacked the large scale KG construction and mining approaches necessary to extract more profound and in-depth insights about the disease and possible cures, treatments, and genetic influences.

NLP techniques have also been utilized outside of the KG construction arena, for example, \cite{tang2020rapidly} introduced CovidQA, a question answering dataset, which comprises 124 questions and answers of triples built by hand from knowledge collected from the CORD-19 dataset. \cite{grujicic2020self} developed a self-supervised context-aware COVID-19 document exploration based on BERT. \cite{liang2020identifying} used BERT to analyze a large collection of COVID-19 literature from the CORD-19 dataset \cite{scite_inc_2020_3724818} to extract COVID-19 related radiological findings. Though rigorous in using large datasets such as CORD-19, the previous NLP techniques were limited in terms of applications and the impact of those applications on the COVID-19 oriented biomedical research field.

\section{Datasets}

\subsection{PubMed dataset}

The PubMed database contains more than 30 million citations within the various fields of life sciences. The PubMed citation database archived by the MEDLINE archive has always been the desired datasets for biomedical text and graph mining research communities. 

We select PubMed dataset because it is a popular dataset in biomedical area and reflect general biomedical knowledge.

\subsection{The PubMed Knowledge Graph}

\cite{xu2020building} built a PubMed KG which connects disambiguated author names, their articles, and bio-entities using the PubMed database were they parsed 29 million PubMed abstracts from 1781 till 2019. In addition to funding extracted from the National Institutes of Health using ExPORTER, and affiliations were extracted from ORCID and MapAffil. 

\subsection{The CORD-19 Dataset}

The CORD-19 dataset was released in response to COVID-19, where the US Government has issued requests for research groups and institutions to combine efforts to release the COVID-19 Open Research Dataset (CORD-19). The datasets contain more than 135000 articles with over 68000 full texts on topics related to Coronavirus and the COVID-19 pandemic. The data set was released to help the biomedical research community by applying the latest in NLP to extract deep insights and understandings of the pandemic patterns and the possible drugs, cures, and genes that might be involved and identified \cite{cord2020}. 

Here we perform our analysis on the entities and relationships extracted from the three datasets and we show the potential in knowledge discovery.





\section{Experiment-0 Identify Experts on Coronavirus topics}
We would like to identify the experts for COVID to encourage collaboration. To do that, we analyzed COVID-19 44k dataset and ranked the researchers according to the number of articles they published in the COVID-19 44K dataset. Part of the results are shown in Table \ref{Related_researchers}. 

\begin{table}[h]
\caption{COVID-19 related researchers}
\centering
\begin{tabular}{l|l}
\hline
Author             & \begin{tabular}[c]{@{}l@{}}\# of articles published\\  in COVID-19 dataset\end{tabular} \\ \hline
Perlman, Stanley   & 142                                                                                     \\ \hline
Drosten, Christian & 137                                                                                     \\ \hline
Yuen, Kwok-Yung    & 136                                                                                     \\ \hline
Baric, Ralph S     & 132                                                                                     \\ \hline
Jiang, Shibo       & 120                                                                                     \\ \hline
Enjuanes, Luis     & 116                                                                                     \\ \hline
Snijder, Eric J    & 104                                                                                     \\ \hline
Weiss, Susan R     & 92  \\ \hline                                         
\end{tabular}
\label{Related_researchers}
\end{table}

\section{Experiment-1 Named Entity Recognition with BioBERT}
\subsection{Model}
BERT (Bidirectional Encoder Representations from Transformers) \cite{devlin2018bert} is a highly influential Natural Language Processing model that proposed back in 2018. BERT was inspired by many advanced Deep Learning models, such as semi-supervised sequence learning\cite{dai2015semi}, ELMo \cite{peters2018deep} and the Transformer architecture \cite{vaswani2017attention}. 

The input representation of BERT is the sum of a token embedding using WordPiece, a segmentation embedding indicating whether each token belongs to sentence A or sentence B, and a position embedding. A [CLS] flag is added before the first word of the sentence, and a [SEP] flag is added as a separator token. 

BERT has two tasks for pre-training: Masked Language Model task and Next Sentence Prediction task. Considering most of traditional NLP model, instead of training a left-to-right or right-to-left model based on the input language, it is better to use the bidirectional model. However, the bidirectional model is not suitable for the conditional task. Thus, inspired by the Cloze task, a masked language model is adapted as the first task for BERT pre-training. The second task for BERT pre-training is Next Sentence Prediction (NSP), which allows the model to understand sentence relationships.  

BioBERT (Bidirectional Encoder Representations from Transformers for Biomedical Text Mining) \cite{lee2020biobert} is a biomedical language representation model based on BERT \cite{devlin2018bert}. It is proposed because directly adapting BERT to text mining in the biomedical area was not promising, given the word shift from generic domain to the biomedical domain. BioBERT is pre-trained on PubMed abstracts and PubMed Central full-text articles (PMC). 

We select BioBERT-Base v1.1 (+PubMed 1M) based on the BERT-base-Cases model. For our fine-tuning section, we fine-tuned it on the NCBI disease dataset. The input to the BioBERT model is a sentence embedded following BERT's embedding process. Parts of the token-level evaluation looks like: "[CLS]Ang \#\#iot \#\#ens \#\#in - converting enzyme 2 (AC \#\#E \#\#2 ) as a SA \#\#RS - Co \#\#V - 2 receptor: molecular mechanisms and potential therapeutic target. [SEP]". The output of this model will be a sentence with labels. Label "B-MISC" means the Begining of bioentities, "I-MISC" means Insdie the bio entities, and "O-MISC" means Outside the bio entity. We tested BioBERT on PubMed KG and the CORD-19 dataset. 

\subsection{Results}
 Examples of entity-level recognized names from the PubMed dataset are shown in Table \ref{table_pubmed_biobert}. The recognized bio-entities are "acute respiratory disease", "pneumonia", and "acute respiratory syndrome coronavirus 2". "SARSCOV-2" should, but is not recognized as bio-entities. 

\begin{table}[t]
\caption{Named Entity Recognition using BioBERT in PubMed dataset}
\centering
\begin{tabular}{ll}
\hline
\textbf{Word}                              & \textbf{Label}  \\ \hline
\multicolumn{1}{l|}{Asymptomatic} & O-MISC \\ \hline
\multicolumn{1}{l|}{carrier}      & O-MISC \\ \hline
\multicolumn{1}{l|}{state}        & O-MISC \\ \hline
\multicolumn{1}{l|}{,}            & O-MISC \\ \hline
\multicolumn{1}{l|}{acute}        & B-MISC \\ \hline
\multicolumn{1}{l|}{respiratory}  & I-MISC \\ \hline
\multicolumn{1}{l|}{disease}      & I-MISC \\ \hline
\multicolumn{1}{l|}{,}            & O-MISC \\ \hline
\multicolumn{1}{l|}{and}          & O-MISC \\ \hline
\multicolumn{1}{l|}{pneumonia}    & B-MISC \\ \hline
\multicolumn{1}{l|}{due}          & O-MISC \\ \hline
\multicolumn{1}{l|}{to}           & O-MISC \\ \hline
\multicolumn{1}{l|}{severe}       & O-MISC \\ \hline
\multicolumn{1}{l|}{acute}        & B-MISC \\ \hline
\multicolumn{1}{l|}{respiratory}  & I-MISC \\ \hline
\multicolumn{1}{l|}{syndrome}     & I-MISC \\ \hline
\multicolumn{1}{l|}{coronavirus}  & I-MISC \\ \hline
\multicolumn{1}{l|}{2}            & I-MISC \\ \hline
\multicolumn{1}{l|}{(}            & O-MISC \\ \hline
\multicolumn{1}{l|}{SARSCoV}      & O-MISC \\ \hline
\multicolumn{1}{l|}{-}            & O-MISC \\ \hline
\multicolumn{1}{l|}{2}            & O-MISC \\ \hline
\multicolumn{1}{l|}{)}            & O-MISC \\ \hline
\multicolumn{1}{l|}{:}            & O-MISC \\ \hline
\multicolumn{1}{l|}{Facts}        & O-MISC \\ \hline
\multicolumn{1}{l|}{and}          & O-MISC \\ \hline
\multicolumn{1}{l|}{myths}        & O-MISC \\ \hline
\end{tabular}
\label{table_pubmed_biobert}
\end{table}

Since we do not have detailed labels for BioBERT fine-tuning training and thus cannot predict detailed labels directly from BioBERT. To get more detailed labels, we trained a Random Forest model on around 100,000 PubMed bio-entities labeled with five categories: species, gene, diseases, drug, gene mutation, and tested on the bio-entities recognized by BioBERT. The F1-score is shown in Table \ref{random_forest}. The F1-Score of disease, gene, and drug recognition are all over 75\%. However, the model poorly predicts when it comes to label "Species" and labels "Gene mutation," probably because the dataset was very unbalanced (Gene mutation only has 17 samples, and Species has only 1395 samples).

\begin{table}[t]
\caption{Specific named entity classification using Random Forest}
\begin{tabular}{l|l|l|l|l}
\hline
\textbf{Label} & \textbf{Precision} & \textbf{Recall} & \textbf{F1-Score} & \textbf{Data Size} \\ \hline
Species        & 0.72               & 0.31            & 0.43              & 15519              \\ \hline
Disease        & 0.94               & 0.61            & 0.74              & 12077               \\ \hline
Gene           & 0.95               & 0.64            & 0.76              & 18678               \\ \hline
Drug           & 0.66               & 0.99            & 0.79              & 53523              \\ \hline
Gene mutation  & 0.5                & 0.17            & 0.25              & 36                 \\ \hline
\end{tabular}
\label{random_forest}
\end{table}

Examples of named entity recognized from CORD-19 44K dataset are shown in Table \ref{table_COVID_biobert}. In the example, the recognized bio-entities are "coronavirus disease 2019", and "COVID-19". "Thrombocytopenia," a kind of disease, should, but is not recognized. From these cases, we find that BioBERT does not perform greatly, despite its high accuracy. It may be because BioBERT can easily recognize the easy and common bio-entities with a high occurrence rate but fail to recognize rare biomedical terms.

\begin{table}[t]
\caption{named entity Recognition using BioBERT in COVID-19 44K dataset}
\centering
\begin{tabular}{ll}
\hline
\textbf{Word}                              & \textbf{Label}  \\ \hline
\multicolumn{1}{l|}{Thrombocytopenia} & O-MISC \\ \hline
\multicolumn{1}{l|}{is}      & O-MISC \\ \hline
\multicolumn{1}{l|}{associated}        & O-MISC \\ \hline
\multicolumn{1}{l|}{with}            & O-MISC \\ \hline
\multicolumn{1}{l|}{severe}        & O-MISC \\ \hline
\multicolumn{1}{l|}{coronavirus}  & B-MISC \\ \hline
\multicolumn{1}{l|}{disease}      & I-MISC \\ \hline
\multicolumn{1}{l|}{2019}            & I-MISC \\ \hline
\multicolumn{1}{l|}{COVID}          & B-MISC \\ \hline
\multicolumn{1}{l|}{-}    & I-MISC \\ \hline
\multicolumn{1}{l|}{19}          & I-MISC \\ \hline
\multicolumn{1}{l|}{infections}           & O-MISC \\ \hline
\multicolumn{1}{l|}{A}       & O-MISC \\ \hline
\multicolumn{1}{l|}{meta}        & O-MISC \\ \hline
\multicolumn{1}{l|}{-}  & O-MISC \\ \hline
\multicolumn{1}{l|}{analysis}     & O-MISC \\ \hline
\end{tabular}
\label{table_COVID_biobert}
\end{table}
\section{Experiment-2.1 Co-occurrence frequency based Knowledge Graph}
\subsection{Method}
We used Gephi to build the co-occurrence frequency based Knowledge Graph. Co-occurrence frequency is an above-chance frequency of occurrence of two entities from an article. The data is from the PubMed Knowledge Graph. For each target node (related bio-entities), we calculated the times it shows up with the source node and treat the times(co-occurrence frequency) as target node's weight. The higher the co-occurrence frequency, the closer the target node is to the source node.

\subsection{Result}
\begin{figure}[t]
    \centering
    \includegraphics [scale=0.4] {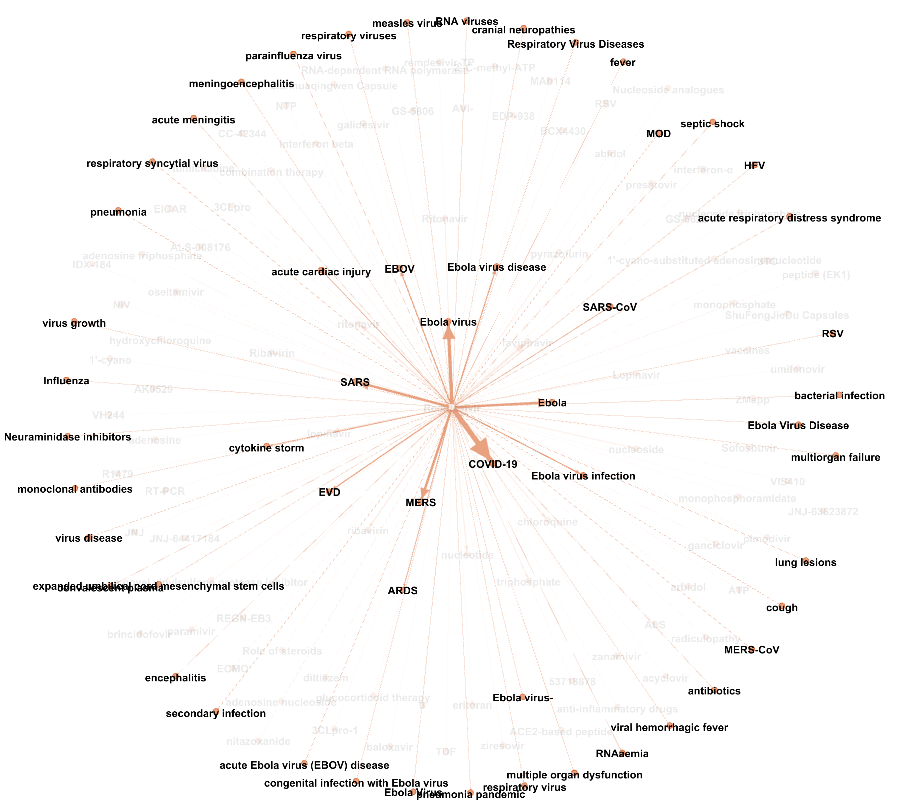}
    \caption{remdesivir related disease KG based on co-occurrence frequency}
    \label{fig:remdesivir_related_disease}
\end{figure}

\begin{figure}
    \centering
    \includegraphics [scale=0.4] {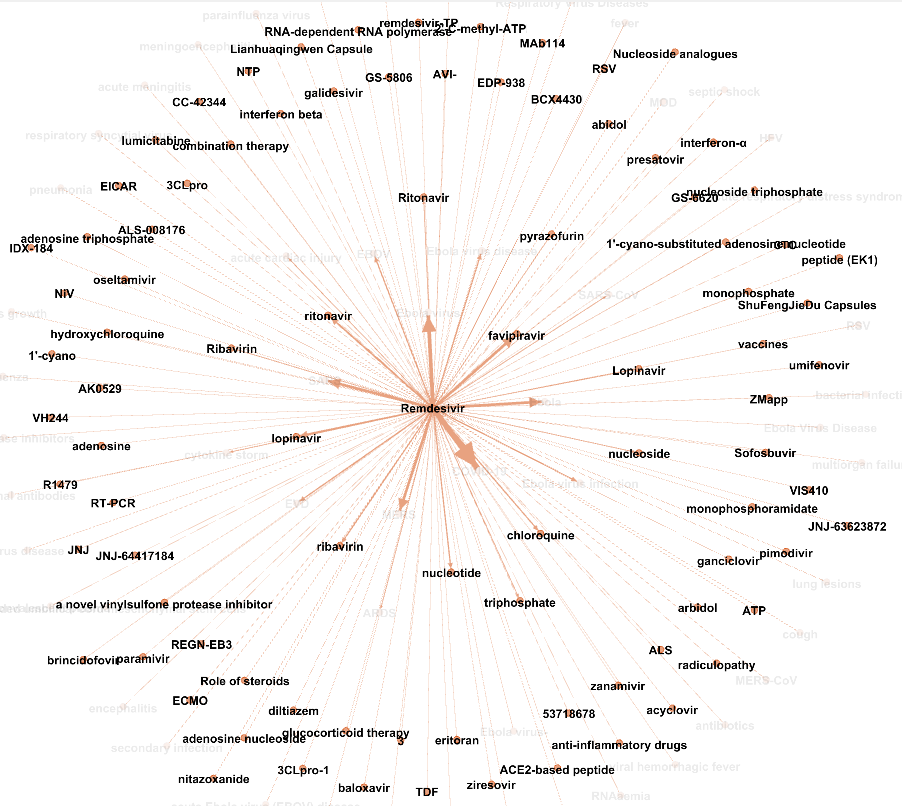}
    \caption{remdesivir related drug KG based on co-occurrence frequency}
    \label{fig:remdesivir_related_drug}
\end{figure}

Figure \ref{fig:remdesivir_related_disease} and Figure \ref{fig:remdesivir_related_drug} shows remdesivir centered KGs. In Figure \ref{fig:remdesivir_related_disease}, the source node is remdesivir and the target nodes are remdesivir related disease. As shown, `` COVID-19", ``Ebola", ``SARS", ``EVD", ``MERS", ``EBOV", ``cytokine storm", ``acute cardiac injury", and `` ARDS" are related to remdesivir. In Figure \ref{fig:remdesivir_related_drug} the source node is remdesivir and the target nodes are remdesivir related drugs from PubMed bio-entity knowledge graph. The edge's weights are based on co-occurrence frequency. As shown, ``favipiravir" (used for influenza), ``ritonavir" (used for HIV), ``lopinavir (used for HIV)", ``ribavirin" (used for severe lung infections), ``chloroquine" (used for lupus, malaria, and rheumatoid arthritis), and ``pyrazofurin", which has antibiotic, antiviral and anti-cancer properties with severe side effects, are remdesivir related drugs.  


We also generated other 6 drug-centered KGs. The 6 drugs are favipiravir, ritonavir, lopinavir, ribavirin, tamiflu, and umifenovir. The diseases highly related to favipiravir are: Ebola, influenza, Epstein-Barr virus, HBV infection, CCHF, avian influenza, hemorrhagic fever, Lassa fever, thrombocytopenia syndrome, Rift Valley fever, and etc.
The drugs highly related to favipiravir are: ribavirin, tamiflu, peramivir, amantadine, laninamivir, BCX4430, T-1105, Pyrazine, and etc.

The diseases highly related to ritonavir are: HIV, AIDS, hepatitis C virus, cirrhosis, nausea, and etc. 
The drugs highly related to ritonavir are:  lopinavir, indinavir, darunavir, lamivudine, atazanavir, nelfinavir, and etc. 

The diseases highly related to lopinavir are: HIV, AIDS, cardiovascular disease, lipodystrophy, hepatitis C, diarrhea, malaria, nausea, and etc. The drugs highly related to lopinavir are: ritonavir, saquinavir, abacavir, nucleoside, indinavir, lamivudine, darunavir, atazanavir, nelfinavir, tenofovir, nevirapine, zidovudine, amprenavir, and etc.

The diseases highly related to ribavirin are: HIV, HCC, cirrhosis, hepatitis, liver cirrhosis, liver transplantation, AIDS, RSV, depression, chronic diseases, and etc.
The drugs highly related to ribavirin are: telaprevir, ledipasvir, alanine, sofosbuvir, boceprevir, PEG, daclatasvir, simeprevir, ritonavir, paritaprevir, ombitasvir, and dasabuvir.

The diseases highly related to tamiflu are: influenza, avian influenza, CCHF, pneumonia, HBV infection, cough, headache, hypernatremia, fever, and etc.  
The drugs highly related to tamiflu are: amantadine, zanamivir, amino acid, oseltamivir, carboxylateribavirin, oseltamivir Phosphate, Peramivir, rimantadine, and laninamivir.

The diseases highly related to umifenovir are: influenza, acute respiratory infections, viral infections, pneumonia, fever, and mepatitis B virus (HBV) infection. The drugs highly related to umifenovir are tamiflu, rimantadine, ribavirin, ingavirin, amantadine, ARB,indole, Zanamivir, Triazavirin, Reaferon, and etc. 

\begin{figure}[t]
    \centering
    \includegraphics [scale=0.33] {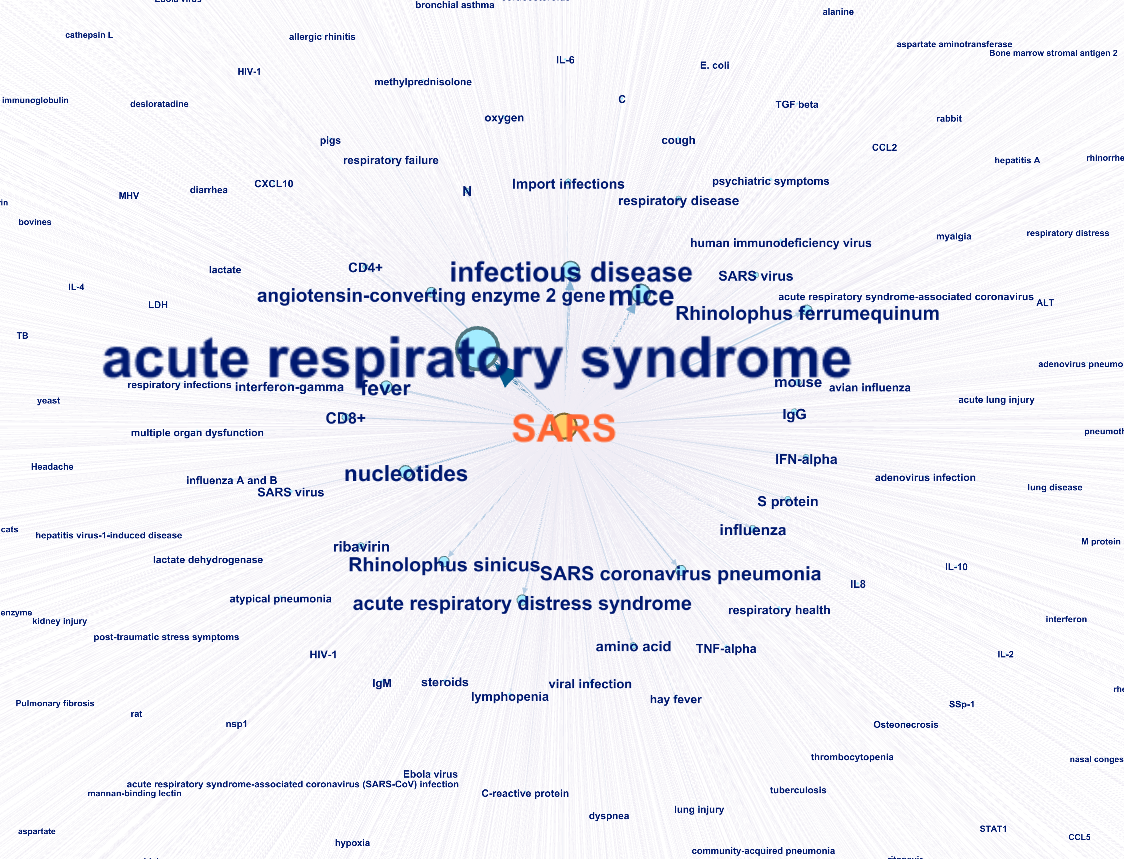}
    \caption{SARS centered KG based on co-occurrence frequency}
    \label{fig:SARS}
\end{figure}

\begin{figure}[t]
    \centering
    \includegraphics [scale=0.3] {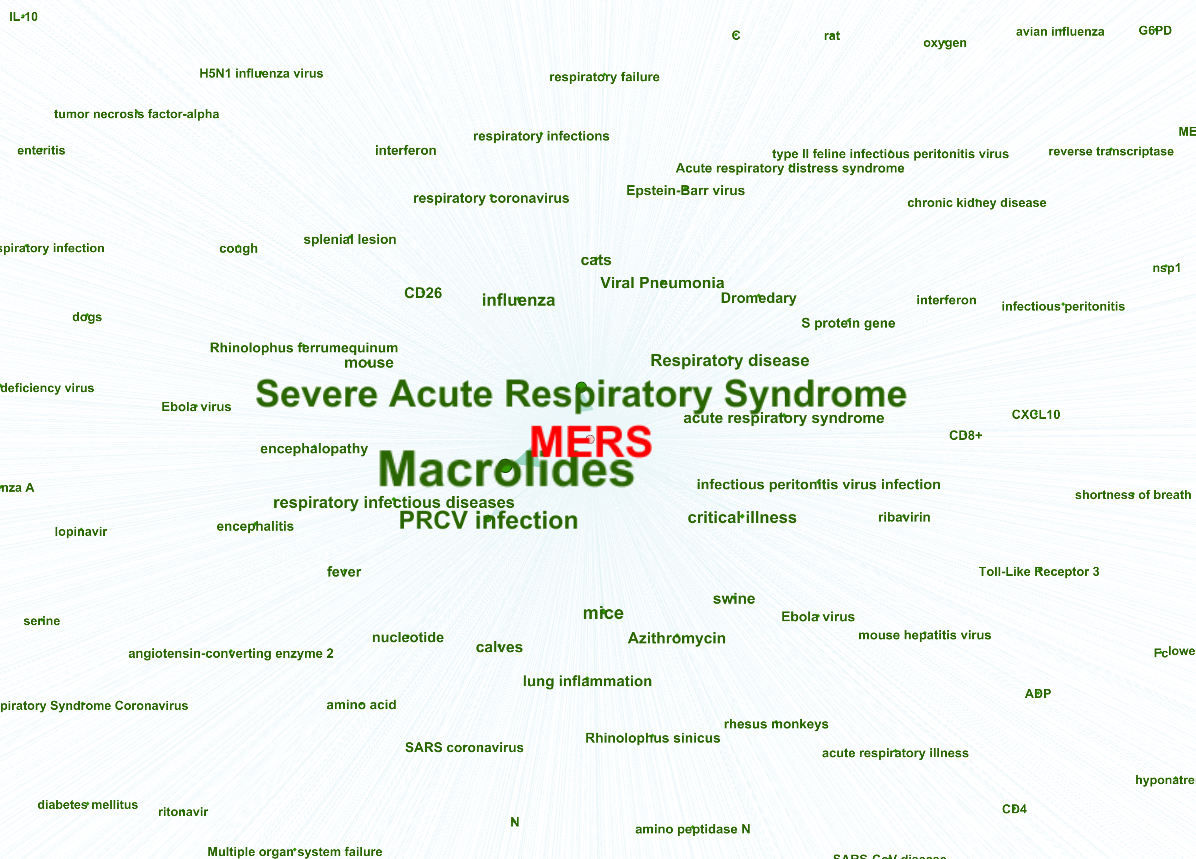}
    \caption{MERS centered KG based on co-occurrence frequency}
    \label{fig:MERS}
\end{figure}

\begin{figure}[t]
    \centering
    \includegraphics [scale=0.32] {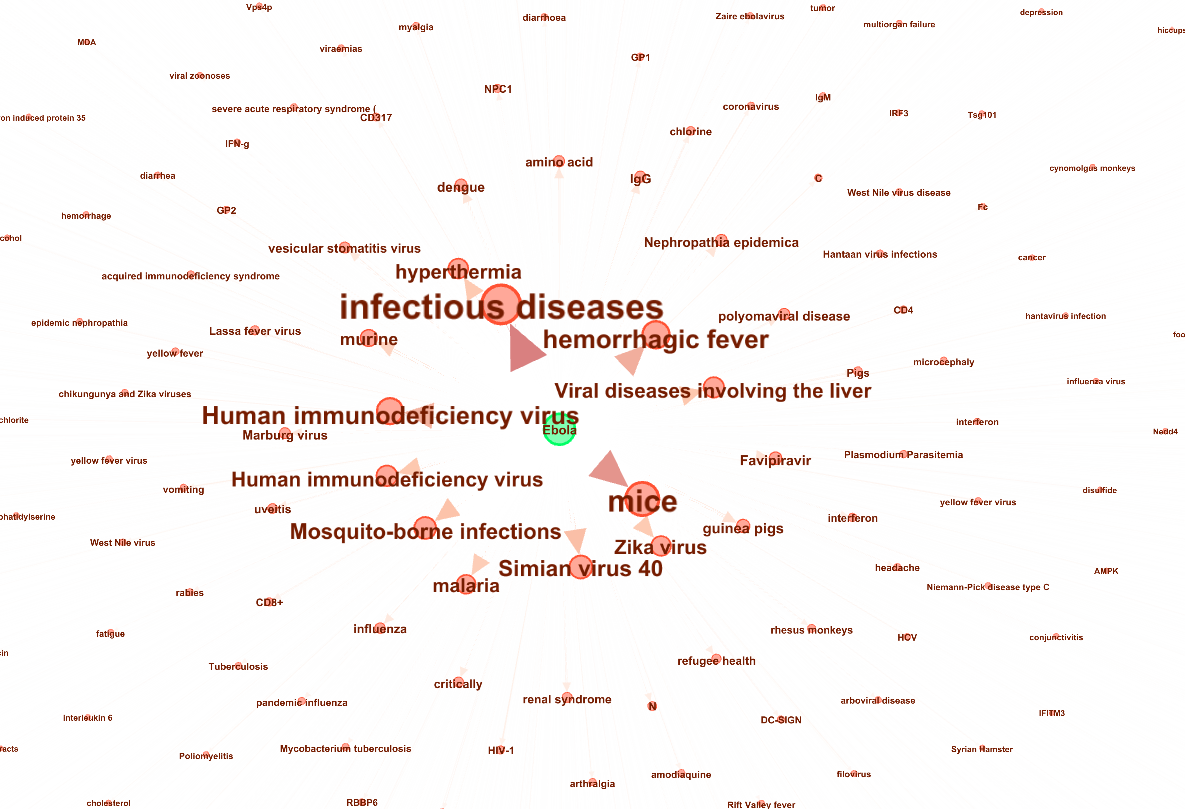}
    \caption{Ebola centered KG based on co-occurrence frequency}
    \label{fig:Ebola}
\end{figure}

Figure \ref{fig:SARS}, Figure \ref{fig:MERS}, and Figure \ref{fig:Ebola} show two corona virus diseases (SARS, MERS) and Ebola centered KG, respectively. Figure \ref{fig:SARS} shows that SARS's highly related diseases are acuate respiratory distress syndrome, fever, influenza, HIV-1, Osteonecrosis, allergic inflammation, lung disease, atypical pneumonia, allergic rhinitis, cough, and etc.
SARS's highly related genes/chemicals are CD8+, CD4+, TNF-alpha, interferon-gamma, IFN-alpha, IL8, lgG, C-reactive protein, S protein, lactate dehydrogenase, ACE2 gene, and etc. SARS's highly related drugs are ribavirin, methylpredinisolone, and corticosteroids, and etc.

Figure \ref{fig:MERS} shows that MERS's highly related diseases are Severe Acute Respiratory Syndrome, PRCV infection, and influenza. MERS's highly related drugs are macrolides, ribavirin, azithromycin, lopinavir, and nitonavir. MERS's highly related genes/chemicals are CD26, S protein gene, amino peptidase N, and etc.

Figure \ref{fig:Ebola} shows that Ebola's highly related diseases are hemorrhagic fever, hyperthermia, malaria, and mosquito-borne infections. MERS's highly related drugs are favipiravir and amodiaquine. MERS's highly related genes/chemicals are CD8+, CD4, DC-SIGN, CD317, GP2, IFN-g, IRF3, RBBP6, and etc.


\begin{figure}[t]
    \centering
    \includegraphics [scale=0.28] {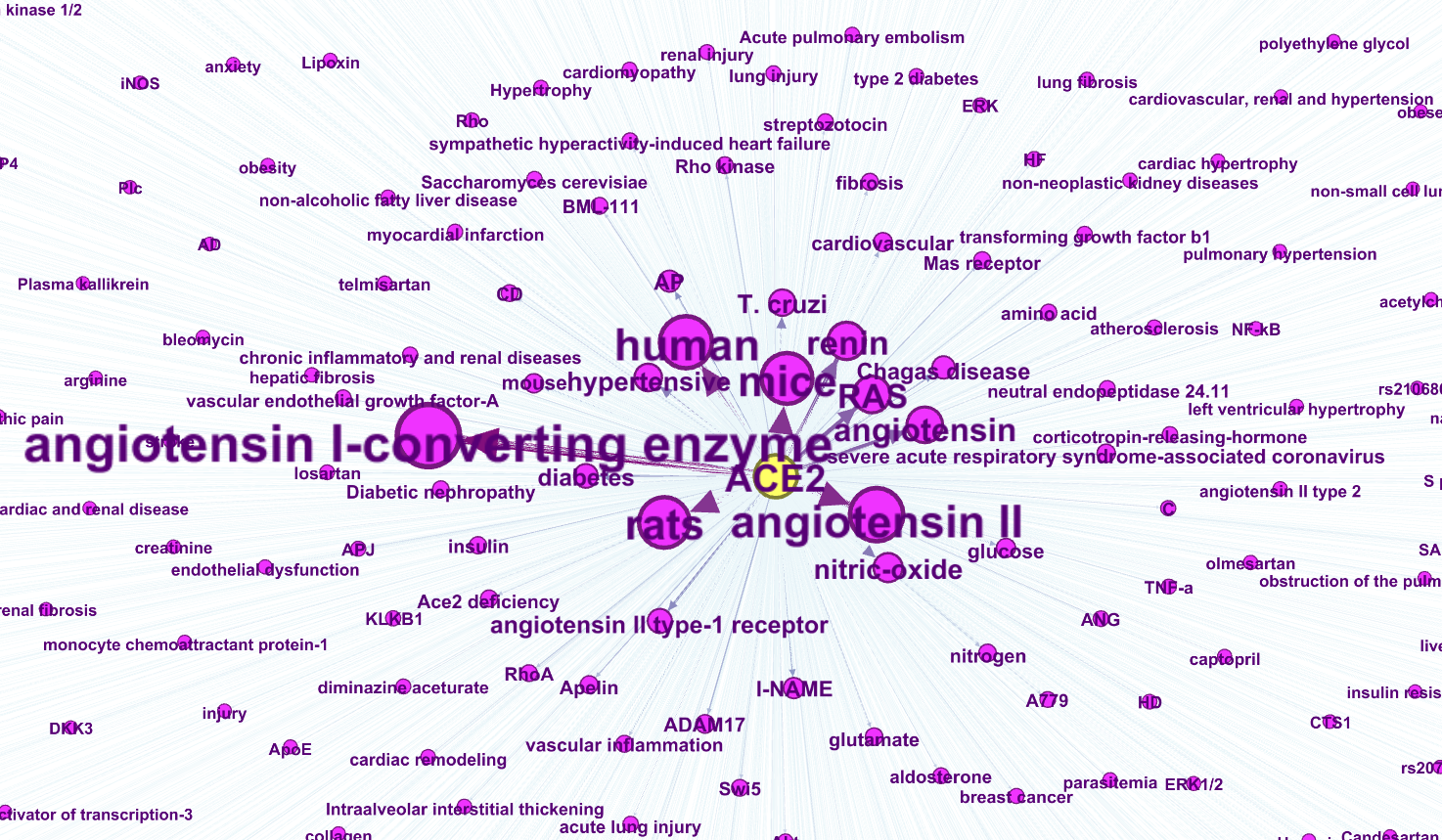}
    \caption{ACE2 centered KG based on co-occurrence frequency}
    \label{fig:ACE2}
\end{figure}

Figure \ref{fig:ACE2} shows Angiotensin-converting enzyme 2 (ACE2) centered Knowledge Graph. ACE2 is an enzyme, which lowers blood pressure by catalysing the hydrolysis of angiotensin II into angiotensin (1-7). ACE2 is the receptor that COVID-19 uses to infect lung cells. It also serves as receptor for other coronaviruses such as HCoV-NL63, SARS-CoV. As shown, ACE2's related genes/chemicals are renin, RAS, angiotensin, insulin, Mas receptor, vascular endothelial growth factor-A, and etc. ACE2's related diseases are diabetes, hypertensive, chagas disease, severe acute respiratory syndrome-associated coronavirus. ACE2's related drugs are streptozotocin, nitric-oxide, and aldosterone .ACE2's related gene mutations are ``rs2106809" and ``rs2074192".

From the results we believe the PubMed Knowledge Graph is very promising. However, this kind of KG has a entity name disambiguation issue. For example, ``Favipiravir" could also be shown as ``favipiravir". Another case is ``ACE-2", which is the abbreviation of Angiotensin-converting enzyme 2. Besides, the co-occurrence frequency cannot reflect the relationship between the source node and the target node well. For example, if ``A has nothing to do with B" mentioned lots of times in different documents, its co-occurrence frequency will be very high.

\section{Experiment-2.2 Cosine similarity based Knowledge Graph}
\subsection{Method}
To deal with problems with co-occurrence frequency based KG, we first normalized the entity using some human designed rules to deal with entity name disambiguation issue. We mainly focus on case sensitive, singular and plural, and disambiguation. For example, ``SIAsNN" will be normalized as "siann" and``respiratory illnesses" will be normalized as "respiratory\underline{ }illness". Then we used Word2Vec to convert the normalized entity to vector with length of 100. We then use Cosine Similarity to measure the similarity between the source node and each target node. The Cosine Similarity is defined as follows:
\begin{equation}
\cos ({\bf S},{\bf T})= {{\bf S} {\bf T} \over \|{\bf S}\| \|{\bf T}\|} = \frac{ \sum_{i=1}^{n}{{\bf S}_i{\bf T}_i} }{ \sqrt{\sum_{i=1}^{n}{({\bf S}_i)^2}} \sqrt{\sum_{i=1}^{n}{({\bf T}_i)^2}} }
\end{equation}

The KG based on cosine similarity is also built using Gephi software.

\subsection{Results}
\begin{figure}[t]
    \centering
    \includegraphics [scale=0.5] {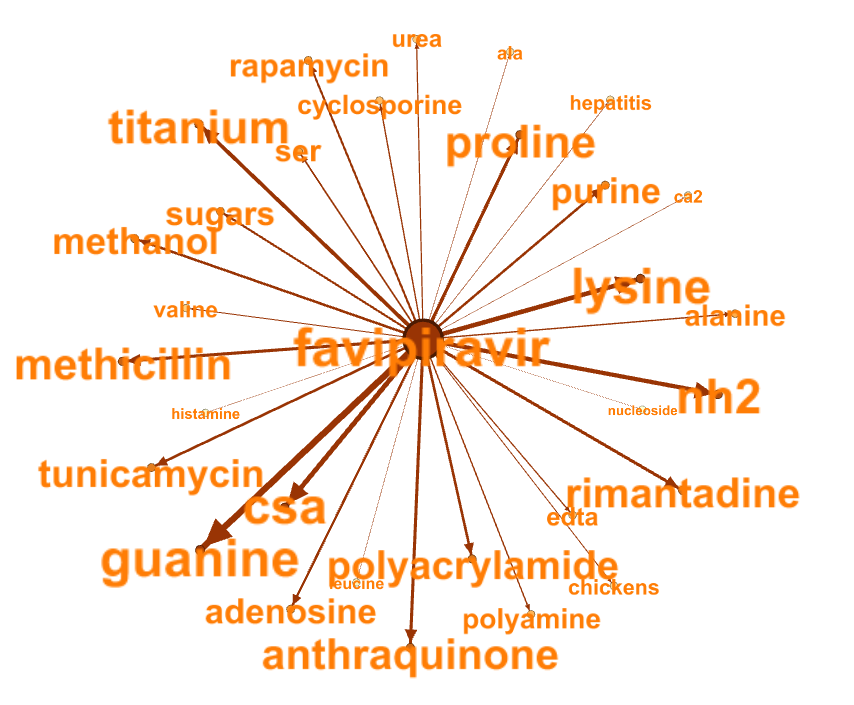}
    \caption{favipiravir-centered KG (related chemical) based on cosine similarity}
    \label{fig:favipiravir_chemical}
\end{figure}

\begin{figure}[t]
    \centering
    \includegraphics [scale=0.5] {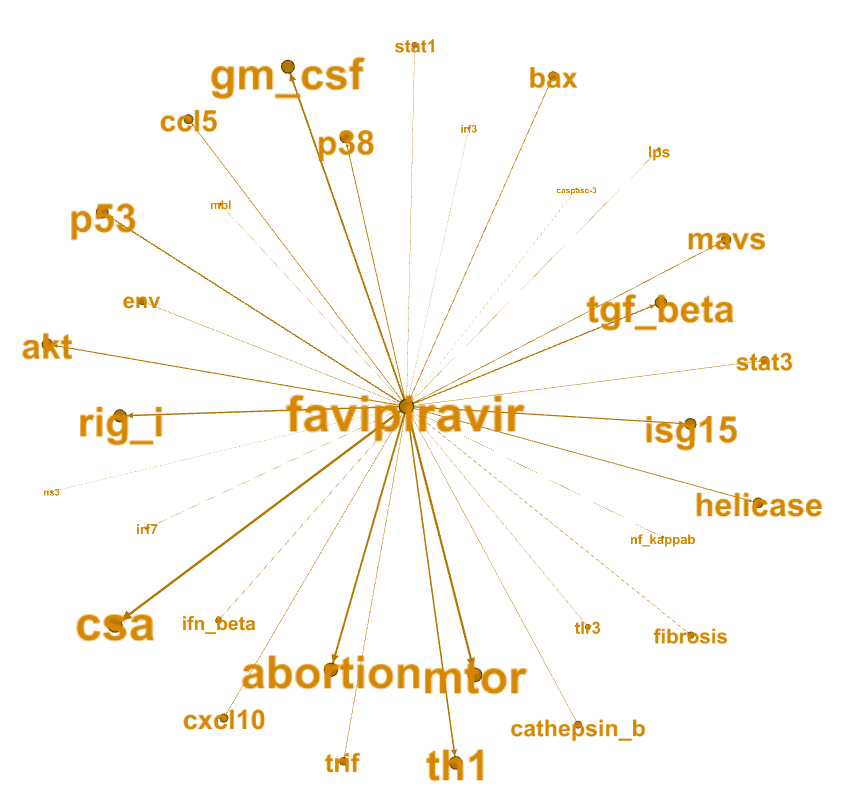}
    \caption{favipiravir-centered KG (related gene) based on cosine similarity}
    \label{fig:favipiravir_gene}
\end{figure}

Figure \ref{fig:favipiravir_chemical} shows part of the favipiravir-centered knowledge graph (chemical related). The source node is favipiravir and the target node are related chemicals. The edge is cosine similarity relations. The closer the target node is to the source node, the similar the target node is to the source node. As shown, the top 10 chemical related to favipiravir are guanine, csa, lysine, nh2, titanium, proline, methicillin, anthraquinone, rimantadine, polyacrylamide. Figure \ref{fig:favipiravir_gene} shows part of the favipiravir-centered knowledge graph (gene related). As shown, the top 10 gene related to favipiravir are gm\underline{~~}csf, abortion, rig\underline{~~}i, isg15, csa, akt, mtor, p53, th1, p38, tgf\underline{~~}beta.

We also generate other 5 drug-centered KGs based on cosine similarity. The top 10 chemicals related to lopinavir are retinoic\underline{~~}acid, nucleoside, tyr, glutamine, ribavirin, glycyrrhizin, co2, lopinavir, phosphonate, lymphoma, ifitm3. The top 10 genes related to lopinavir are neuraminidase, p53, eif2alpha, apod, ribavirin, infection, cox-2, ifitm3, iron.

The top 10 chemicals related to ribavirin are glycyrrhizin, lactate, corticosteroid, coronavirus, steroid, nucleoside, ribavirin, sodium, glucose, infection, oxygen, calcium, obesity. The top 10 genes related to ribavirin are toxicity, p53, swine, fibrosis, iron, neuraminidase, diabetes, ribavirin, anemia, inflammation, infection.

The top 10 chemicals related to ritonavir are atp, ritonavir, cyclophosphamide, mtt, toxicity, sialic\underline{~~}acid, sds, encephalitis, superoxide, sucrose, ethanol. The top 10 genes related to ritonavir are jnk, p53, rig\underline{~~}i, encephalitis, rnase\underline{~~}l, stat3, toxicity, akt, neuraminidase, stat1.

The top 10 chemicals related to tamiflu are superoxide, prednisolone, flavonol, proline, nitric\underline{~~}oxide, thymidine, glycyrrhizin, propidium\underline{~~}iodide, nitrogen, aspirin, tamiflu. The top 10 genes related to tamiflu are il-10, cd44, eif2alpha, tgf\underline{~~}beta1, tlr2, ifn, cxcl10, tumor\underline{~~}necrosis\underline{~~}factor\underline{~~}tnf)-alpha, ire1, ccl2, tbk1.

The top 10 chemicals related to umifenovir are: tacrolimus, alkyl, carbon\underline{~~}monoxide, ca(2, cd, nucleolin, cytosine, glycyrrhizic\underline{~~}acid, 2'-o, umifenovir, prostaglandin\underline{~~}e2. The top 10 genes related to umifenovir are parp, pd\underline{~~}l1, monocyte\underline{~~}chemoattractant\underline{~~}protein-1, nef, cxcr4, cd45, nucleolin, dc\underline{~~}ign, annexin\underline{~~}v, cd19, mmp-2.

\section{Discussion and Conclusion}




In this research, we first used BioBERT to recognize entities in the PubMed and the CORD-19 dataset. Our results show that most of the recognized entities are strictly biomedical. Most of the recognized entities in the CORD-19 dataset are disease lacking diversity in entity types due to a lack in finding a suitable bio-medical training dataset with detailed labeled bio-entity. For future work, we will explore more other bio-medical dataset and try other biomedical NLP models for named entity recognition, e.g., blueBERT \cite{peng2019transfer}. 

Furthermore, we introduced the construction of Coronavirus Knowledge Graph based on two different methods: co-occurence frequency and cosine similarity. We explored and revealed that the drug candidates recommended by drug-centered KG are promising. We will consult experts in COVID-related research to verify our drug-centered KG and conduct a more in-depth analysis for future work.

Also, we aim to build a wider COVID related KG, connecting all COVID related bio-entities rather than small drug/disease-centered KG. The extracted KG will help understand the relationships between the diseases, the genes, the viruses, and the cures involved in and related to COVID-19. 

\begin{figure}[t]
    \centering
    \includegraphics [scale=0.35] {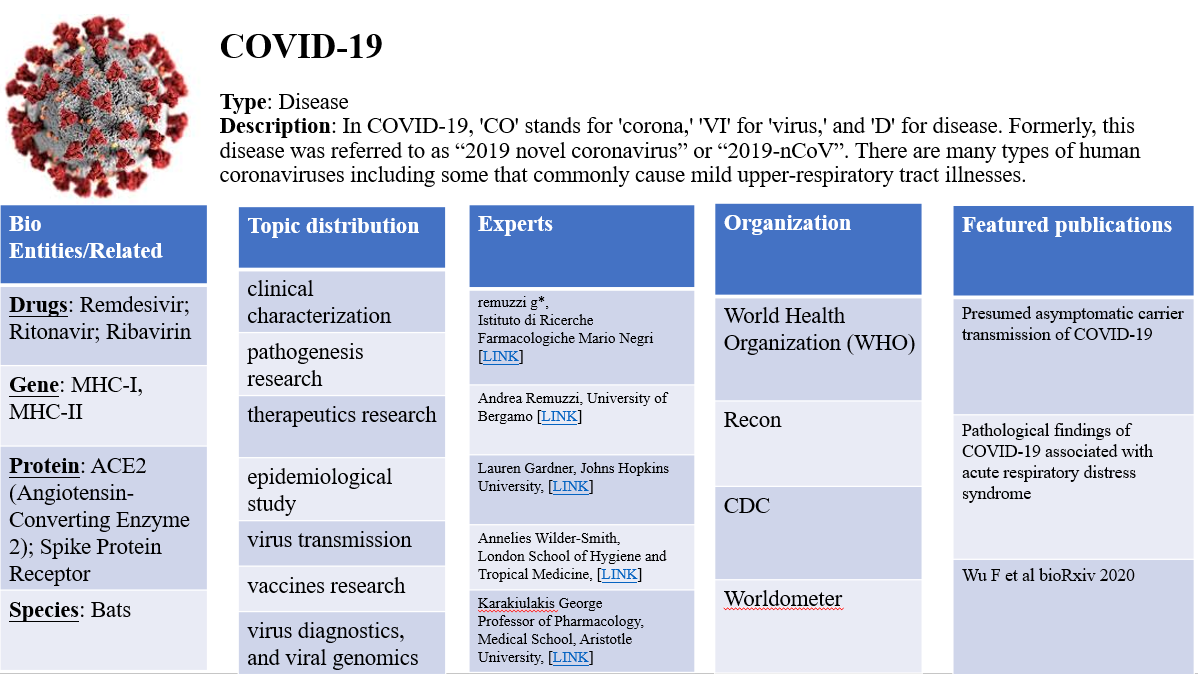}
    \caption{auto drug profiling example}
    \label{fig:profiling}
\end{figure}

Finally, we hope to build an automatic profiling system to generate expert, drug, or disease profiling. The expected disease profiling will look like Figure \ref{fig:profiling}, which includes description, related bio entities (drugs, gene, protein, species), topic distribution, related experts, organization, and featured publications. 

\section{Contributions}
Y.D. and Y.B. proposed the idea and supervised the project. C.C. wrote the paper. I.A.E wrote the Introduction and revised this paper. C.C. conducted the named entity recognition and Knowledge Graph building. I.A.E conducted the Word2Vec for Experiment 2.2.
\begin{acks}
We would like express our gratitude to Prof. Jaewoo Kange's DMIS Lab team for pretraining BioBERT, Vinay Locharulu for suggestion and support, Prof. Jian Xu for providing PubMed Knowledge Graph, and Yifei Wu for conducting entity normalization.
\end{acks}

\bibliographystyle{ACM-Reference-Format}
\bibliography{sample-base}











\end{document}